\pdfoutput=1
\documentclass[11pt]{article}
\usepackage{ACL2023}
\usepackage{times}
\usepackage{latexsym}
\usepackage{arydshln}
\usepackage[T1]{fontenc}
\usepackage[utf8]{inputenc}
\usepackage{microtype}

\usepackage{inconsolata}
\usepackage{hyperref}       
\usepackage{url}            
\usepackage{booktabs}       
\usepackage{amsfonts}       
\usepackage{nicefrac}       
\usepackage{microtype}      
\usepackage{xcolor}         
\usepackage{graphicx}

\usepackage{caption}
\usepackage{subcaption}
\usepackage{wrapfig}
\usepackage{multirow}
\usepackage{amsmath}
\usepackage{amsfonts}
\usepackage{bm}
\usepackage{amssymb}
\usepackage{amsthm}
\usepackage{verbatim}
\usepackage{longtable}
\usepackage{arydshln}
\usepackage{algorithm}
\usepackage{algpseudocode}
\usepackage{mathtools, nccmath}

\newcommand{\ours}{\texttt{TBT}}

\DeclareMathOperator*{\argmin}{arg\,min}
\DeclarePairedDelimiter{\nint}\lfloor\rceil

\newcommand*\samethanks[1][\value{footnote}]{\footnotemark[#1]}

\title{Binary and Ternary Natural Language Generation}

\author{Zechun Liu\thanks{\hspace{.06in}Equal contribution}  \\
  Reality Labs, Meta Inc. \\
  \texttt{zechunliu@meta.com} \\ \And
  Barlas Oğuz\samethanks \\
  Meta AI \\
  \texttt{barlaso@meta.com} \\ \And
  Aasish Pappu \\
  Meta AI \\
  \texttt{aasish@fb.com} \\ \AND
  Yangyang Shi \\
  Reality Labs, Meta Inc. \\
   \texttt{yyshi@meta.com} \\ \And
  Raghuraman Krishnamoorthi \\
  Reality Labs, Meta Inc. \\
   \texttt{raghuraman@meta.com} \\
  }

\begin{document}
\maketitle
\begin{abstract}
Ternary and binary neural networks enable multiplication-free computation and promise multiple orders of magnitude efficiency gains over full-precision networks if implemented on specialized hardware.  However, since both the parameter and the output space are highly discretized, such networks have proven very difficult to optimize.  The difficulties are compounded for the class of transformer text generation models due to the sensitivity of the attention operation to quantization and the noise-compounding effects of autoregressive decoding in the high-cardinality output space.  We approach the problem with a mix of statistics-based quantization for the weights and elastic quantization of the activations and demonstrate the first ternary and binary transformer models on the downstream tasks of summarization and machine translation.  Our ternary BART base achieves an R1 score of 41 on the CNN/DailyMail benchmark, which is merely 3.9 points behind the full model while being 16x more efficient.  Our binary model, while less accurate, achieves a highly non-trivial score of 35.6.  For machine translation, we achieved BLEU scores of 21.7 and 17.6 on the WMT16 En-Ro benchmark, compared with a full precision mBART model score of 26.8.  We also compare our approach in the 8-bit activation setting, where our ternary and even binary weight models can match or outperform the best existing 8-bit weight models in the literature.  Our code and models are available at: \url{https://github.com/facebookresearch/Ternary_Binary_Transformer}.
\end{abstract}

\section{Introduction}
Generative pre-trained transformers~\cite{brown2020language, lewis2020bart, radford2018improving} have emerged as powerful and generic tools, driving breakthroughs not only in language understanding but the field of AI in general. These models owe their success mainly to their seemingly infinite ability to scale to ever-larger data and model sizes.  Unfortunately, such scaling comes at the cost of large computational requirements, putting extensively large generative transformers out of reach of all but the most resource-rich institutions.  Even moderately sized pre-trained transformers have limited applications due to their size and computational cost.  Making generative transformers more efficient is imperative for widening their use to more devices and practical applications.

In this work, we explore making generative pre-trained transformers more efficient via the quantization of their weights and activations.  Quantizing the weights of a neural network is useful for compression and allows the model to be stored more efficiently.  However, compression alone does not reduce computation costs since the network's activations need to be computed in full precision.  Quantizing both weights and activations allows computation to be performed with lower precision, potentially leading to significant efficiency gains depending on the quantization level and hardware implementation.  Quantizing neural networks have a long history, and multiple works have attempted to quantize pre-trained transformers at various quantization levels ~\cite{shen2020q, TernaryBERT, liu2022bit, qin2021bibert}.  Most of this work focuses on encoder-only models (mainly BERT) for sentence and token classification tasks.  Quantizing text generation models has generally been regarded as a more difficult task~\cite{behnke2021efficient, tao2022compression} due to the large output vocabulary and sequential decoding.  Recent work has tackled this problem, though only for mild quantization levels (down to 8-bit activations) and with mixed success.

In contrast, we are interested in very low-bit quantization, down to ternary and even binary weights and activations. In order to achieve this, we combine and unify best practices for weight and activation quantization and present a framework that uses gradient-matching quantization for weights and elastic quantization for activations. We apply our method to natural language generation tasks and, for the first time, demonstrate low-bit generative transformers of competitive accuracy. Our ternary (weight and activation) model lags a full-precision BART~\cite{lewis2020bart} model by only 4 points in ROUGE on the XSUM summarization dataset. In contrast, our model with ternary weights and 8-bit activations comes within 1 point and even outperforms comparable state-of-the-art models with 8-bit weights. We also demonstrate a fully binary (weights and activations) model. While not as competitive, it is able to achieve a highly non-trivial ROUGE-1 score of 31.7.

Our results also extend to machine translation models. On the WMT16 En-Ro benchmark, we quantize an mBART model to extend the ternary-weight 8-bit activation SoTA by 1.2 points while demonstrating fully ternary and fully binary translation models for the first time.

We summarize our contributions as follows:

\noindent $\bullet$ We propose a novel combination of statistics-based weight quantization with learning-based activation quantization, which enables stably training transformer encoder-decoder models to converge in the fully ternary/binary settings, which was not previously possible.

\noindent $\bullet$ We significantly improve the state-of-the-art text generation models in the 8-bit activation and ternary/binary weight settings while setting the first non-trivial baselines for the fully ternary and fully binary settings.

\section{Method}
In this section, we first introduce the previous practices in binarization and ternarization. Then, we introduce a unified statistic-based weight binarization / ternarization method that can alleviate the gradient mismatch issue and enhance the quantized weights entropy. Lastly, we analyze the difference between weight quantization and activation quantization and propose an elastic ternarization method for activations. We abbreviate our method as $\ours{}$, short for ``\textbf{T}ernary / \textbf{B}inary \textbf{T}ransformer''.

\subsection{Preliminary}
\subsubsection{Ternarization}
Ternary neural networks, where real values are quantized to three levels, are first introduced in~\cite{li2016ternary}. Thus, these values can be represented in 2 bits, leading to a 16$\times$ reduction in size and computation. Moreover, the computations can be calculated multiplication-free, leading to even further computation gains on suitable hardware. The recent work integrates the ternarization algorithm in natural language models for quantizing the weights and activations in classification tasks ~\cite{TernaryBERT} and ternarizing the weight (8-bit activations are used) in generative models~\cite{li2022dq,tao2022compression}. The general formula~\cite{li2016ternary} for ternarization is as follows:

\begin{equation}
\label{eq:prev_ternary}
\!\!\!\!  \mathbf{X}_\mathbf{T}^i =\! \left\{
         \begin{array}{lr}
         \vspace{0.3em}
         \!\! - \alpha_{_\mathbf{T}}, & {\rm if} \ \mathbf{X}_\mathbf{R}^i < -\Delta \\
         \vspace{0.3em}
         \!\! 0, &  {\rm if} \ -\Delta \leqslant \mathbf{X}_\mathbf{R}^i \leqslant \Delta \\
         \!\! + \alpha_{_\mathbf{T}}, & {\rm if} \ \mathbf{X}_\mathbf{R}^i > \Delta \\
         \end{array}
         \right.
\end{equation}
\begin{equation}
\Delta = \frac{0.7 \cdot ||\mathbf{X_R}||_{l1}}{n_{_{\mathbf{X_R}}}}
\end{equation}
\begin{equation}
\alpha_{_\mathbf{T}} = \frac{\sum_i \mathbf{X}_\mathbf{R}^i\cdot\mathbf{1}_{| \mathbf{X}_\mathbf{R}^i|>\Delta}}{\sum_i \mathbf{1}_{| \mathbf{X}_\mathbf{R}^i|>\Delta}}
\end{equation}
Here $\mathbf{X_T}$ denotes the ternary weights/activations, and $\mathbf{X_R}$ represents their real-valued counterparts.  $n_{_{\mathbf{X_R}}}$ denotes the total number of elements in the tensor. $\Delta$ is the ternary threshold, and $\alpha_{_\mathbf{T}}$ is the scaling factor that minimizes l2-loss between $\mathbf{X_T}$ and $\mathbf{X_R}$.

\subsubsection{Binarization}
The neural network binarization denotes representing the weights and/or activation with bi-level values. It is first proposed in BNN~\cite{courbariaux2016binarized} and has evolved in the follow-up works~\cite{rastegari2016xnor,liu2018bi}. \citet{rastegari2016xnor} formulates binarization as:
\begin{equation}
\label{eq:prev_binary}
\!\!\!\!  \mathbf{X}_\mathbf{B}^i = \alpha_{_\mathbf{B}} \! \cdot \! {\rm Sign}(\mathbf{X}_\mathbf{R}^i) =\! \left\{
         \begin{array}{lr}
         \vspace{0.3em}
         \!\! - \alpha_{_\mathbf{B}}, & \!\!\! {\rm if} \ \mathbf{X}_\mathbf{R}^i < 0 \\
         \!\! + \alpha_{_\mathbf{B}}, & \!\!\! {\rm if} \ \mathbf{X}_\mathbf{R}^i \geqslant 0
         \end{array}
         \right.
\end{equation}
\begin{equation}
\alpha_{_\mathbf{B}}  = \frac{||\mathbf{X_R}||_{l1}}{n_{_{\mathbf{X_R}}}}
\end{equation}
Here $\mathbf{X_B}$ can represent binary weights or binary activations. $\alpha_{_\mathbf{B}}$ denotes the scaling-factor that minimize the l2 loss between $\mathbf{X}_\mathbf{R}$ and $\alpha_{_\mathbf{B}} \! \cdot \! {\rm Sign}(\mathbf{X}_\mathbf{R})$.

The acceleration and compression effect of ternary/binary neural networks is significant. By representing the weights and activations with $\{-1, 0, 1\}$, the network enjoys $\sim$16$\times$ memory saving compared to its 32-bit floating-point counterpart. When further binarize the weights and activations to only 1-bit (i.e., $\{-1, 1\}$), up to 32$\times$ model-size reduction and 58$\times$ speedup on CPUs have been achieved ~\cite{rastegari2016xnor}, where the matrix multiplication operations are replaced with light-weighted bitwise XNOR operations.

Despite its appealing characteristics, naively binarizing or ternarizing the transformer model for natural language generation results in several accuracy drops or even a total failure in training.  It has been observed that the attention layers of the transformer network are difficult to quantize to low bits.  Also, the auto-regressive decoding tends to accumulate errors due to quantization.  Given the nature of generative language networks that require high-precision output, quantizing both the activations and weights in these models to extreme bit values is non-trivial and has not been explored before.

\begin{figure*}[t!]
    \centering
    \includegraphics[width=0.8\linewidth]{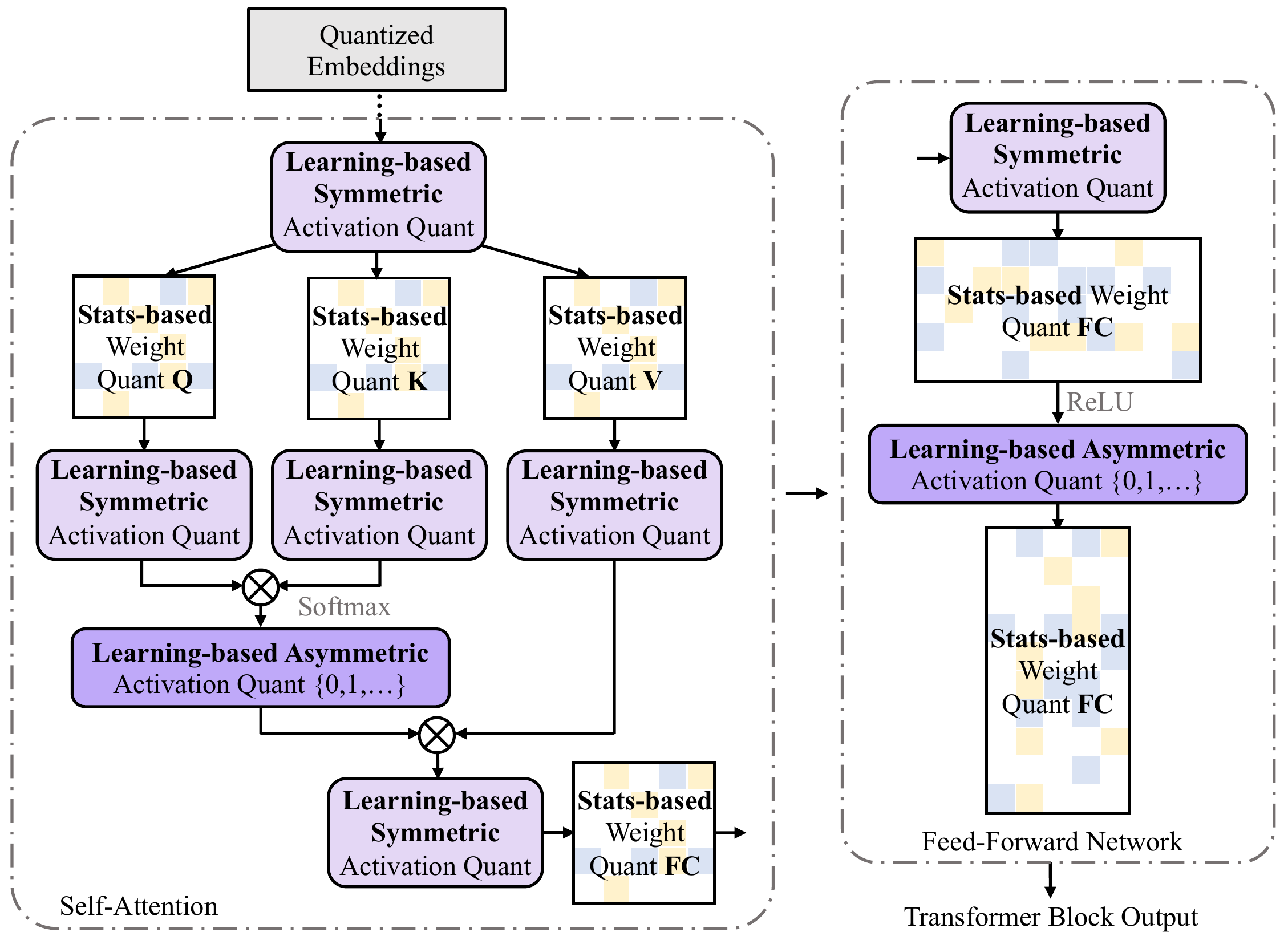}
    \caption{Overview of \ours{}. A transformer block contains the multi-head self-attention and feed-forward network. We propose a statistic-based quantization method for weights ternarization/binarization and adopt a learning-based asymmetric quantization method for activation in ReLU/Softmax output ($\mathbf{X} \in \mathbb{R}^n_+$) and learning-based asymmetric quantization method for activations that contain both positive and negative values in other layers ($\mathbf{X} \in \mathbb{R}^n$).}
    \label{fig:overview}
\end{figure*}

\subsection{Stats-based max-entropy isometric weight quantization}
\label{sec:stats_weight}
We propose a statistics-based method for weight binarization/ternarization. Particularly, this novel quantization method considers maximizing the entropy of the quantized weights and reducing the gradient mismatch in the backward pass. Previous works~\cite{courbariaux2016binarized, bai2021binarybert, TernaryBERT} are mainly focused on minimizing the $l2$ loss between the quantized weights and the real-valued weights to find the optimal quantization scheme,
\begin{equation}
\label{eq:l2_loss}
    \alpha^* = \argmin || \alpha \hat{\mathbf{W}}_\mathbf{Q} - \mathbf{W}_\mathbf{R} ||_{l2}
\end{equation}
where $\hat{\mathbf{W}}_\mathbf{Q}$ denotes binary/ternary weights and $\alpha^*$ denotes the optimal scaling factor calculated.
Despite the broad application and great success of the classic quantization scheme, we found that merely minimizing the $l2$ loss neglects several critical but intractable issues in ultra-low-bit weight quantization: (1) The information entropy of the quantized weights is not considered. Eq.~\ref{eq:prev_ternary} and Eq.~\ref{eq:prev_binary} calculate the quantized weights to minimize the distance to the real-valued weights, which could lead to imbalanced quantized weight distribution and harm the quantized weights representation capacity. (2) The quantization function Eq.~\ref{eq:prev_ternary} and Eq.~\ref{eq:prev_binary} are not isometric, meaning that it does not consider the magnitude consistency between the quantized weights and real-valued weights, while we find that magnitude consistency contributes significantly to accurate gradient estimation.

Considering the above two limitations in previous solutions, we are motivated to design a novel quantization function that enhances information entropy and reduces gradient mismatch. To boost the weights representation capability, in information theory, more information is preserved when the quantized weights contain higher entropy:
\begin{equation}
\label{eq:entropy}
\max_{p_i} \ \mathcal{H} = -p_i \log(p_i), s.t. \sum_{i=1}^{N} p_i = 1
\end{equation}
with $p_i$ denoting the proportion of real-valued weights being quantized to $i^{th}$ quantization level in total $N$ levels. Eq.~\ref{eq:entropy} can be easily solved with a Lagrange multiplier, and the optimal $p_i^* = \frac{1}{N}, \ \ i \in \{1,2,…, N\}$, suggesting the best quantization scheme to preserve maximum information entropy is to distribute the real-valued weights in all quantization levels as evenly as possible.

For reducing the gradient mismatch, as suggested by the previous binarization work~\cite{liu2020bi}, the magnitude difference between the quantized weight and the real-valued weight will greatly influence the gradient scale and a mismatch in magnitude will be amplified in back-propagation and cause gradient vanishing or explosion during training. Thus it is important to ensure the magnitude of real-valued weights and quantized weights are consistent.

Combining two requirements discussed above, we proposed max-entropy isometric weight quantization.
In ternarization, it is formulated as
\begin{equation}
\label{eq:ternary_weight}
\begin{split}
    &\mathbf{W}_\mathbf{T}^i = \alpha_{_{\mathbf{T}}}  \nint{{\rm Clip}(\frac{\mathbf{W}_\mathbf{R}^i - \mu_{_{\mathbf{T}}}}{\alpha_{_{\mathbf{T}}}}, -1, 1)}  \\
    & {\rm where}  \ \ \mu_{_{\mathbf{T}}} = \overline{\mathbf{W}_\mathbf{R}}, \\
    & \ \ \ \ \ \ \ \ \ \ \ \ \alpha_{_{\mathbf{T}}} = \frac{4}{3} \cdot \frac{||\mathbf{W}_\mathbf{R} - \mu_{_{\mathbf{T}}}||_{l1}}{n_{_{\mathbf{W}_\mathbf{R}}}}
\end{split}
\end{equation}
Where $\mathbf{W}_\mathbf{T}$ and $\mathbf{W}_\mathbf{R}$ refer to the ternary weights and real-valued weights, respectively. The rounding function $\nint{\cdot}$ and $\rm{Clip}(\cdot)$ function quantize weights to $\{-1, 0 ,1\}$. $\mu_{_{\mathbf{T}}}$ is the mean of real-valued weights and $n_{_{\mathbf{W}_\mathbf{R}}}$ denotes the number of weights in the weight matrix. Scaling factor $\alpha$ is calculated from the weight statistics and follows the entropy rule to scale the real-valued weight $\mathbf{W}_\mathbf{R}$ to be evenly distributed in quantization levels. In the ternary case, the weights are quantized to $\{-\alpha_{_{\mathbf{T}}}, 0, \alpha_{_{\mathbf{T}}}\}$. When the real-valued weights are initialized as uniformly and symmetrically distributed~\cite{he2015delving, glorot2010understanding}, the scaling factor $\alpha_{_{\mathbf{T}}}$ will distribute $\frac{\mathbf{W}_\mathbf{R}^i}{\alpha_{_{\mathbf{T}}}}$ to $[-1.5, 1.5]$, such that the output ternary weights will have near uniform distribution in three ternary levels. Meanwhile, Eq.~\ref{eq:ternary_weight} is an isometric mapping where the real-valued weights are scaled by $\frac{1}{\alpha_{_{\mathbf{T}}}}$ to near [-1, 1] and time $\alpha_{_{\mathbf{T}}}$ to scale back after quantization. In this way, the magnitude is preserved.

Correspondingly, in the binary case we have,
\begin{equation}
\label{eq:binary_weight}
\begin{split}
    & \mathbf{W}_\mathbf{B}^i = \alpha_{_{\mathbf{B}}} \cdot \rm{Sign}(\frac{\mathbf{W}_\mathbf{R}^i - \mu_{_{\mathbf{B}}} }{\alpha_{_{\mathbf{B}}}}) \\
    & {\rm where}  \ \ \mu_{_{\mathbf{B}}} = \overline{\mathbf{W}_\mathbf{R}}, \\
    & \ \ \ \ \ \ \ \ \ \ \ \ \alpha_{_{\mathbf{B}}} = \frac{||\mathbf{W}_\mathbf{R} - \mu_{_{\mathbf{B}}}||_{l1}}{n_{_{\mathbf{W}_\mathbf{R}}}}
\end{split}
\end{equation}
Here $\mathbf{W}_\mathbf{B}$ denotes the binary weights, where substracting the average $\mu_{_{\mathbf{B}}}$ makes the real-valued weight zero-centered before binarization and thus encourages an even distribution in binarized weights. Then the scaling factor $\alpha_{_{\mathbf{B}}}$ matches the magnitude between real-valued and binary weights. Particularly, in Eq.~\ref{eq:binary_weight}, $\mathbf{W}_\mathbf{B}^i = \alpha_{_{\mathbf{B}}} \cdot \rm{Sign}(\frac{\mathbf{W}_\mathbf{R}^i - \mu_{_{\mathbf{B}}} }{\alpha_{_{\mathbf{B}}}}) = \alpha_{_{\mathbf{B}}} \cdot \rm{Sign}(\mathbf{W}_\mathbf{R}^i - \mu_{_{\mathbf{B}}})$, we explicitly include the $\alpha_{_{\mathbf{B}}}$ in the denominator to keep the binarization function isometric and the gradients \textit{w.r.t.} weights can be calculated straightforwardly as:
\begin{equation}
\label{eq:binary_weight_backward}
\begin{split}
    \frac{\partial \mathbf{W}_\mathbf{B}^i}{\partial \mathbf{W}_\mathbf{R}^i} \overset{STE}{\approx} \mathbf{1}_{|\frac{\mathbf{W}_\mathbf{R}^i - \mu_{_{\mathbf{B}}} }{\alpha_{_{\mathbf{B}}}}|<1}
\end{split}
\end{equation}
STE is abbreviated for straight-through estimator~\cite{bengio2013estimating}, which replaces the non-differentiable ${\rm Sign}$ function with ${\rm Clip}$ function in the backward pass. We show that the proposed max-entropy isometric weight quantization improves the accuracy of weight binarization / ternarization by $6.0$ / $11.53$  RougeL scores on the CNN/DailyMail benchmark, respectively. More details can be found in Sec.~\ref{sec:experiment_summarization}.

\subsection{Learning-based activation quantization}
In contrast to neural network weights that are stored on the disk, activations are calculated on-the-fly. The distribution of activations in a particular layer depends on the network weights as well as the corresponding input sequence, and thus varies from batch to batch. In order to have the quantization function better capture the underlying activation distribution, we propose learning-based activation quantization.

Inspired by BiT~\cite{liu2022bit}, we divide the activation layers into two categories: the activation layers with non-negative values ($\mathbf{X}_\mathbf{R} \! \in \! \mathbb{R}_+$), \textit{i.e.}, Softmax/ReLU layer outputs and the rest of the layers with both positive and negative activations ($\mathbf{X}_\mathbf{R} \! \in \! \mathbb{R}$). We binarize / ternarize the first activation category ($\mathbf{X}_\mathbf{R} \! \in \! \mathbb{R}_+$) to $\{0, \alpha\}$ / $\{0, \alpha, 2\alpha \}$, and symmetrically quantize the later activation category ($\mathbf{X}_\mathbf{R} \! \in \! \mathbb{R}$) to $\{-\alpha, \alpha\}$ and $\{-\alpha, 0, \alpha\}$ in binary and ternary cases respectively. In this way, the activation distribution matches the original full-precision activations and thus reduces the quantization error. Further, we learn to scale the real-valued activations to better fit quantization thresholds, and this learnable scaling factor can be updated end-to-end with the gradients from the network loss to better account for overall network optimization.

In the ternary case, we propose the elastic ternarization function formulated as,
\begin{equation}
\begin{split}
\label{eq:ternary_activation}
\!\!\!\mathbf{X}_\mathbf{T}^i & = \alpha_{_\mathbf{T}} \mathbf{\hat{X}_T}^i  \\
   & = \!\!\left\{
         \begin{array}{lr}
         \vspace{0.3em}
         \!\!\!\alpha_{_\mathbf{T}} \nint{{\rm Clip}(\frac{\mathbf{X}_\mathbf{R}^i}{\alpha_{_\mathbf{T}}}, 0, 2)}, \ \  {\rm if} \ \mathbf{X}_\mathbf{R} \! \in \! \mathbb{R}_+ \\
         \!\!\!\alpha_{_\mathbf{T}} \nint{{\rm Clip}(\frac{\mathbf{X}_\mathbf{R}'^i}{\alpha_{_\mathbf{T}}}, -\!1, 1)},  {\rm if} \  \mathbf{X}_\mathbf{R} \! \in \! \mathbb{R}
         \end{array}
         \right.
\end{split}
\end{equation}
where $\mathbf{X}_\mathbf{R}$ and $\mathbf{X}_\mathbf{T}$ denote real-valued and ternary activations, respectively. To keep the formula concise, we set $\mathbf{X}_\mathbf{R}' = \mathbf{X}_\mathbf{R} - \overline{\mathbf{X}_\mathbf{R}}$, denoting the zero-mean real-valued activations. $\alpha_{_\mathbf{T}}$ is the scaling factor.
Different from the weight quantization, the scaling factor in Eq.~\ref{eq:ternary_activation} is learned with the gradient update. We follow the practice in~\cite{zhou2016dorefa, esser2019learned} to calculate the gradients with straight-through estimation (STE) bypassing the non-differentiable rounding function:
\begin{equation}
\begin{split}
\label{eq:ternary_activation_back}
\!\!\!\!\! \frac{\partial\mathbf{X}_\mathbf{T}^i}{\partial \alpha_{_\mathbf{T}}} \! & \overset{STE}{\approx} \\
\!\!\! & \! \left\{
     \begin{array}{lr}
     \vspace{0.3em}
     \!\!\!\mathbf{\hat{X}_T}^i -\! \frac{\mathbf{X}_\mathbf{R}^i}{\alpha_{_\mathbf{T}}} \! \cdot \! \mathbf{\large{1}}_{0 \leqslant \mathbf{X}_\mathbf{R}^i \leqslant 2\alpha_{_\mathbf{T}}}, {\rm if} \ \mathbf{X}_\mathbf{R} \! \in \! \mathbb{R}_+ \\
     \!\!\!\mathbf{\hat{X}_T}^i -\! \frac{\mathbf{X}_\mathbf{R}'^i}{\alpha_{_\mathbf{T}}} \! \cdot \! \mathbf{\large{1}}_{|\mathbf{X}_\mathbf{R}'^i| \leqslant \alpha_{_\mathbf{T}}}, \ \ \ \ {\rm if} \ \mathbf{X}_\mathbf{R}\! \in \! \mathbb{R}
     \end{array}
     \right.
\end{split}
\end{equation}
The learnable scaling factor can dynamically adapt to different activation distributions and improve the ternarization accuracy. In the binary case, it is formulated as.

\begin{equation}
\begin{split}
\label{eq:binary_activation}
\!\!\!\mathbf{X}_\mathbf{B}^i & = \alpha_{_\mathbf{B}} \mathbf{\hat{X}_B}^i  \\
   & =\!\left\{
         \begin{array}{lr}
         \vspace{0.5em}
         \!\!\alpha_{_\mathbf{B}} \nint{{\rm Clip}(\frac{\mathbf{X}_\mathbf{R}^i}{\alpha_{_\mathbf{B}}}, 0, 1)}, \ {\rm if} \ \mathbf{X}_\mathbf{R}\!\in\!\mathbb{R}_+ \\
         \!\!\alpha_{_\mathbf{B}} \cdot {\rm Sign}(\frac{\mathbf{X}_\mathbf{R}'^i}{\alpha_{_\mathbf{B}}}), \ \ \ \ \ \ \ \ \ {\rm if} \ \mathbf{X}_\mathbf{R}\!\in\!\mathbb{R}
         \end{array}
         \right.
\end{split}
\end{equation}
Here $\mathbf{X}_\mathbf{B}$ denotes the binary activations.

Correspondingly, the gradients \textit{w.r.t.} the scaling factor $\alpha$ can be easily calculated as
\begin{equation}
\begin{split}
\label{eq:binary_activation_back}
\!\!\! \frac{\partial\mathbf{X}_\mathbf{B}^i}{\partial \alpha_{_\mathbf{B}}} & \overset{STE}{\approx} \\
& \!\!\! \left\{
     \begin{array}{lr}
     \vspace{0.3em}
     \!\!\mathbf{\hat{X}_B}^i \!-\! \frac{\mathbf{X}_\mathbf{R}^i}{\alpha_{_\mathbf{B}}} \!\cdot\! \mathbf{\large{1}}_{0 \leqslant \mathbf{X}_\mathbf{R}^i \leqslant \alpha_{_\mathbf{B}}}, \ {\rm if} \ \mathbf{X}_\mathbf{R} \! \in \! \mathbb{R}_+ \\
     \!\!{\rm Sign}(\mathbf{X}_\mathbf{R}'^i), \ \ \ \ \ \ \ \ \ \ \ \ \ \ \ \ \ \ \ \ \! {\rm if} \ \mathbf{X}_\mathbf{R}\! \in \! \mathbb{R}
     \end{array}
     \right.
\end{split}
\end{equation}
We demonstrate that with the learning-based activation quantization method and statistics-based weight quantization scheme, the proposed $\ours{}$ for the first time is able to quantize the BART model for natural language generation tasks to ternary and even binary weights and activations, and achieve reasonable accuracy on summarization and translation benchmarks.

\section{Experiments}
In this section, we evaluate the effectiveness of our low-bit quantization scheme for natural language generative model on text summarization benchmarks: CNN/DailyMail~\cite{nallapati2016abstractive} and XSUM~\cite{narayan-etal-2018-dont}. We additionally experiment on the machine translation task with mBART on WMT16 English-Romanian (En-Ro) dataset~\cite{bojar-etal-2016-findings}.

\subsection{Experimental settings}
We follow recent work~\cite{li2022dq} in training the quantized network with initialization and knowledge distillation from a full-precision pre-trained model. Specifically, we use the BART-base~\cite{lewis2019bart} as our full-precision baseline for summarization tasks and mBART-large~\cite{liu2020multilingual} for the translation task. We train the quantized models for 20 epochs on 8 GPUs with a batch size of 128 and a learning rate of 2.5e-4 for 8-bit activation models and 5e-4 for binary and ternary activation models.

\begin{table*}[t]
\renewcommand\arraystretch{0.6}
\centering
\caption{Comparison of quantization methods for text summarization on XSUM and CNN/DailyMail benchmarks. We use the “E-W-A (\#bits)” notation referring to the number of bits of embeddings, weights and activations, (specifically, $1$ denotes binary, $2$ denotes ternary). The results of QuantBart, DQ-BART and BlockPruning are quoted from their paper. Additionally, we implement the algorithm developed in BinaryBert, BiBert and TernaryBert to the BART model and report the results, denoted with $^*$. We use the rouge-$\{$1,2,L$\}$ as evaluation metrics.}
\label{tab:main}
\setlength{\tabcolsep}{1.4mm}
\vspace{-0.5em}
{\resizebox{0.95\textwidth}{!}{
\begin{tabular}{lcccccccccc}
\hline
\noalign{\smallskip}
& & & & \multicolumn{3}{c}{\textbf{XSUM}} & \multicolumn{3}{c}{\textbf{CNN/DailyMail}} \\
\textbf{Method} & \begin{tabular}[c]{@{}c@{}}\textbf{\#Bits}$_\text{ (E-W-A)}$\end{tabular}& \begin{tabular}[c]{@{}c@{}}\textbf{Size}$_\text{ (MB)}$\end{tabular} & \begin{tabular}[c]{@{}c@{}}\textbf{FLOPs}\end{tabular} & R1 & R2 & RL & R1 & R2 & RL \\ 
\noalign{\smallskip}\hline\noalign{\smallskip}
BART & 32-32-32 & 532.0 & 1$\times$ & 43.84 & 20.79 & 35.71 & 44.90 & 22.25 & 42.09 \\ 
\noalign{\smallskip}
QuantBart~\cite{tao2022compression} & 8 - 8 - 8 & 138.1 & -- & 40.25 & 17.78 & 32.70 & -- & -- & -- \\
\noalign{\smallskip}
DQ-BART~\cite{li2022dq} & 8 - 8 - 8 & 138.1 & -- & 42.51 & 19.61 & 34.61 & 44.66 & 21.92 & 41.86 \\
\noalign{\smallskip}\hline\noalign{\smallskip}
\multicolumn{3}{l}{\textit{Ternary}} & \\
\noalign{\smallskip}\hdashline[0.8pt/1pt]\noalign{\smallskip}
Baseline (TWN)~\cite{li2016ternary} & 2 - 2 - 8 & 39.6 & 0.25$\times$ & 39.99 & 17.13 & 31.99 & 42.99 & 20.05 & 40.18\\
\noalign{\smallskip}
QuantBart~\cite{tao2022compression} & 2 - 2 - 8 & 39.6 & 0.25$\times$ & 39.15 & 16.72 & 31.72 & -- & -- & -- \\
\noalign{\smallskip}
DQ-BART~\cite{li2022dq} & 2 - 2 - 8 & 39.6 & 0.25$\times$ & 40.06 & 17.34 & 32.46 & 42.94 & 20.07 & 40.13 \\
\noalign{\smallskip}
\textbf{\ours{}} & 2 - 2 - 8 & 39.6 & 0.25$\times$ & \textbf{42.40} & \textbf{19.54} & \textbf{34.51} & \textbf{43.46} & \textbf{20.52} & \textbf{40.58} \\
\noalign{\smallskip}\hdashline[0.8pt/1pt]\noalign{\smallskip}
Baseline (TWN)~\cite{li2016ternary} & 2 - 2 - 2 & 39.6 & 0.0625$\times$ & 12.80 & 1.21 & 11.4 & 12.92 & 0.32 & 12.42\\
\noalign{\smallskip}
TernaryBert$^*$~\cite{TernaryBERT} & 2 - 2 - 2 & 39.6 & 0.0625$\times$ & 14.03 & 2.23 & 11.79 & 10.95 & 0.52 & 8.56 \\
\noalign{\smallskip}
\textbf{\ours{}} & 2 - 2 - 2 & 39.6 & 0.0625$\times$ & \textbf{36.21} & \textbf{14.38} & \textbf{29.07} & \textbf{41.03} & \textbf{18.18} & \textbf{38.30} \\
\noalign{\smallskip}\hline\noalign{\smallskip}
\multicolumn{3}{l}{\textit{Binary}} & \\
\noalign{\smallskip}\hdashline[0.8pt/1pt]\noalign{\smallskip}
Baseline (BWN)~\cite{courbariaux2016binarized} & 1 - 1 - 8 & 23.2 & 0.125$\times$ & 1.90 & 0.01 & 1.78 & 2.78 & 0.08 & 2.48\\
\noalign{\smallskip}
BinaryBert$^*$~\cite{bai2021binarybert} & 1 - 1 - 8  & 23.2 & 0.125$\times$ & 39.76 & 17.05 & 31.99 & 40.66 & 18.52 & 28.36 \\
\noalign{\smallskip}
BlockPruning ~\cite{lagunas2021block} & -- & 23 & -- &  -- & -- & -- & 41.4 & 18.7 & 38.4\\
\noalign{\smallskip}
\textbf{\ours{}} & 1 - 1 - 8 & 23.2 & 0.125$\times$ & \textbf{40.96} & \textbf{18.37} & \textbf{33.30} & \textbf{42.66} & \textbf{19.72} & \textbf{39.80} \\
\noalign{\smallskip}\hdashline[0.8pt/1pt]\noalign{\smallskip}
Baseline (BWN)~\cite{courbariaux2016binarized} & 1 - 1 - 1 & 23.2 & 0.0156$\times$ & 1.90 & 0.01 & 1.78 & 2.78 & 0.08 & 2.48\\
\noalign{\smallskip}
BinaryBert$^*$~\cite{bai2021binarybert}  & 1 - 1 - 1 & 23.2 & 0.0156$\times$ & 8.13 & 0.12 & 7.69 & 9.80 & 0.15 & 8.62 \\
\noalign{\smallskip}
BiBert$^*$~\cite{qin2021bibert} & 1 - 1 - 1 & 23.2 & 0.0156$\times$ & 7.58 & 0.06 & 7.54 & 14.22 & 0.13 & 10.06 \\
\noalign{\smallskip}
\textbf{\ours{}} & 1 - 1 - 1 & 23.2 & 0.0156$\times$ & \textbf{31.68} & \textbf{11.19} & \textbf{25.29} & \textbf{35.56} & \textbf{11.71} & \textbf{33.23} \\
\noalign{\smallskip}
\hline
\end{tabular}}}
\end{table*}

\subsection{Summarization}
\label{sec:experiment_summarization}
For the summarization task, we adopt the following benchmarks:
\paragraph{The XSUM dataset~\cite{narayan-etal-2018-dont}} consists of 226k documents sampled from the online news website of BBC, together with short, one sentence summaries.  Since the summaries are very short, abstractive methods tend to do better on this dataset.
\paragraph{CNN/DailyMail~\cite{nallapati2016abstractive}} is another news summarization benchmark, with longer documents (\textasciitilde 30 sentences) and longer, multi-sentence summaries.  The dataset contains close to 300k document-summary pairs.

\vspace{0.8em}
We use BART-base model~\citep{lewis2019bart}, which is an English-only encoder-decoder transformer with 140 million parameters.  We compare using the standard ROUGE-\{1,2,l\} metrics for this task.

For the ternary weights and 8-bit activations setting, we compare with two state-of-the-art methods QuantBart~\cite{tao2022compression} and DQ-BART~\cite{li2022dq}.  For the fully ternary setting, and the binary quantization experiments, there is no prior art.  Therefore we provide a naive quantization baseline, using popular implementations from previous work~\citep{li2016ternary, courbariaux2016binarized}, and adapt the binary and ternary methods proposed for the BERT models~\cite{bai2021binarybert,qin2021bibert,TernaryBERT} to BART.

Our main results are summarized in Table \ref{tab:main}.  In the ternary weights and 8-bit activations setting, \ours{} improves previous SoTA by up to \textbf{2.3 points} in ROUGE score on XSUM, and up to \textbf{0.5 points} on CNN/DailyMail.  Both improvements are significant.

Further quantizing weights to \emph{binary}, while keeping activations at 8-bit, we are still able to achieve a ROUGE-L score of 33.3 on XSUM, which is 0.8 points higher than the previous \emph{ternary} SoTA (DQ-BART), and comparable on CNN/DailyMail.  This is the first demonstration of a binary-weight generative transformer model of competitive accuracy to our knowledge.  Additionally, \ours{} binary weight BART model achieves \textbf{1.2 points} higher ROUGE score on CNN compared with the SoTA pruning method with the same compressed model size.

Moving on to ternary and binary activations, there is no prior art, and previous implementations fail to produce meaningful results.  Our method, on the other hand, achieves ROUGE-L scores of 29.1 and 38.3 on XSUM and CNN/DailyMail in the fully ternary setting, which are 6.6 and 3.8 points behind the full-precision baseline respectively.  Our fully binary (weights and activations) model has a wider gap at 10.4 and 8.9 points, however still manages to produce highly non-trivial output at ROUGE-L scores of 25.3 and 33.2 points for XSUM and CNN/DailyMail.

\begin{table}[t]
\renewcommand\arraystretch{0.6}
\centering
\caption{Comparison of quantization methods on mBART-large model for translation on WMT16 En-Ro.}
\label{tab:translation}
\setlength{\tabcolsep}{1.4mm}
\vspace{-0.8em}
{\resizebox{0.48\textwidth}{!}{
\begin{tabular}{llcccccccccccc}
\hline
\noalign{\smallskip}
\textbf{Method} & \begin{tabular}[c]{@{}c@{}}\textbf{\#Bits}$_\text{ (E-W-A)}$\end{tabular}& \begin{tabular}[c]{@{}c@{}}\textbf{Size}$_\text{ (GB)}$\end{tabular} & \textbf{BLEU} \\
\noalign{\smallskip}\hline\noalign{\smallskip}
mBART~\cite{liu2020multilingual} & 32-32-32 & 2.44 & 26.82 \\
\noalign{\smallskip}
DQ-BART~\cite{li2022dq} & 8 - 8 - 8  & 0.61 & 25.91 \\
\noalign{\smallskip}
DQ-BART~\cite{li2022dq} & 2 - 2 - 8  & 0.31 & 23.48 \\
\noalign{\smallskip}\hline\noalign{\smallskip}
\textbf{\ours} & 2 - 2 - 8  & 0.31 & \textbf{24.63}\\
\noalign{\smallskip}
\textbf{\ours} & 2 - 2 - 2  & 0.31 & \textbf{21.70} \\
\noalign{\smallskip}
\textbf{\ours} & 1 - 1 - 8  & 0.16 & \textbf{24.30}\\
\noalign{\smallskip}
\textbf{\ours} & 1 - 1 - 1  & 0.16 & \textbf{17.59}\\
\noalign{\smallskip}\hline
\end{tabular}}}
\end{table}

\subsection{Machine translation}
\label{sec:experiment_translation}
We also evaluate our model on machine translation.  We adopt the En-Ro benchmark from the WMT'16 shared task~\citep{bojar2016results} to be compatible with previous work.  Our base model is an mBART-large model~\cite{liu2020multilingual}, a 680 million parameter multi-lingual encoder-decoder transformer pre-trained on 25 languages.

Table \ref{tab:translation} shows our results.  In the ternary weight setting with 8-bit activations, we improve the previous SoTA by 1.2 points, achieving 24.63 BLEU.  Remarkably our binary weight model also outperforms the previous ternary weight SoTA by almost a full point.  It scores 24.3 BLEU -- only 1.5 points behind a full mBART model while being 16$\times$ smaller.

In the fully ternary and binary settings, where previous methods failed to converge, \ours{} models are able to reach practical levels of performance, with ternary \ours{} mBART achieving 21.7 BLEU, and \ours{} binary mBART at 17.59.

\begin{table}[t]
\renewcommand\arraystretch{0.6}
\centering
\caption{Ablation study on the effects of the proposed learning-based activation quantization method and stats-based weight quantization method on XSUM and CNN/DailyMail benchmark.}
\vspace{-0.8em}
\label{tab:ablation}
\setlength{\tabcolsep}{1.4mm}
{\resizebox{0.48\textwidth}{!}{
\begin{tabular}{llcccccccccccc}
\hline
\noalign{\smallskip}
& & & \multicolumn{3}{c}{\textbf{XSUM}} \\
& \textbf{Method} & \begin{tabular}[c]{@{}c@{}}\textbf{\!\!\!\!\#Bits}$_\text{ (E-W-A)}$\end{tabular}& R1 & R2 & RL \\
\noalign{\smallskip}\hline\noalign{\smallskip}
\small{1}\!\!\! & Baseline (TWN) & 2 - 2 - 2 & 12.80 & 1.21 & 11.4 \\
\noalign{\smallskip}
\small{2}\!\!\! & + Activation(learning-based) & 2 - 2 - 2 & 15.05 & 1.38 & 12.13 \\
\noalign{\smallskip}
\small{3}\!\!\! & + Weight(stats-based) & 2 - 2 - 2 & 13.79 & 0.87 & 12.74 \\ 
\noalign{\smallskip}
\small{4}\!\!\! & + Both & 2 - 2 - 2 & \textbf{36.21} & \textbf{14.38} & \textbf{29.07} \\
\noalign{\smallskip}\hdashline[0.8pt/1pt]\noalign{\smallskip}
\small{5}\!\!\! & Baseline (BWN) & 1 - 1 - 1 & 1.90 & 0.01 & 1.78 \\
\noalign{\smallskip}
\small{6}\!\!\! & + Activation(learning-based) & 1 - 1 - 1 & 1.90 & 0.01 & 1.78 \\
\noalign{\smallskip}
\small{7}\!\!\! & + Weight(stats-based) & 1 - 1 - 1 & 10.96 & 0.29 & 10.00  \\ 
\noalign{\smallskip}
\small{8}\!\!\! & + Both & 1 - 1 - 1 & \textbf{31.68} & \textbf{11.19} & \textbf{25.29} \\
\noalign{\smallskip}\hline\noalign{\smallskip}
& & & \multicolumn{3}{c}{\textbf{CNN/DailyMail}} \\
& & & R1 & R2 & RL \\
\noalign{\smallskip}\hline\noalign{\smallskip}
\ \small{9}\!\!\! & Baseline (TWN) & 2 - 2 - 2 & 12.92 & 0.32 & 12.42 \\
\noalign{\smallskip}
\small{10}\!\!\! & + Activation(learning-based) & 2 - 2 - 2  & 13.34 & 0.99 & 12.58 \\
\noalign{\smallskip}
\small{11}\!\!\! & + Weight(stats-based) & 2 - 2 - 2 & 19.34 & 0.42 & 18.42 \\ 
\noalign{\smallskip}
\small{12}\!\!\! & + Both & 2 - 2 - 2 & \textbf{41.03} & \textbf{18.18} & \textbf{38.30} \\
\noalign{\smallskip}\hdashline[0.8pt/1pt]\noalign{\smallskip}
\small{13}\!\!\! & Baseline (BWN) & 1 - 1 - 1 & 2.78 & 0.08 & 2.48 \\
\noalign{\smallskip}
\small{14}\!\!\! & + Activation(learning-based) & 1 - 1 - 1 & 2.78 & 0.08 & 2.48 \\
\noalign{\smallskip}
\small{15}\!\!\! & + Weight(stats-based) & 1 - 1 - 1 & 15.05 & 0.35 & 14.01 \\ 
\noalign{\smallskip}
\small{16}\!\!\! & + Both & 1 - 1 - 1 & \textbf{35.56} & \textbf{11.71} & \textbf{33.23} \\
\noalign{\smallskip}\hline
\end{tabular}}}
\end{table}

\begin{table}[t]
\renewcommand\arraystretch{0.6}
\centering
\caption{Generated average sequence length comparison between baseline method and our method.}
\vspace{-0.8em}
\label{tab:length}
\setlength{\tabcolsep}{3mm}
{\resizebox{0.45\textwidth}{!}{
\begin{tabular}{llcccccccccccc}
\hline
\noalign{\smallskip}
\textbf{Method} & \begin{tabular}[c]{@{}c@{}}\textbf{\#Bits}$_\text{ (E-W-A)}$\end{tabular}& \textbf{XSUM} & \textbf{CNN/DailyMail} \\
\noalign{\smallskip}\hline\noalign{\smallskip}
BART-base & 32-32-32 & 30.73 & 99.89 \\
\noalign{\smallskip}\hline\noalign{\smallskip}
Baseline & 2 - 2 - 8 & 28.53 & 93.63\\
\noalign{\smallskip}
\textbf{\ours} & 2 - 2 - 8  & 32.04 & 95.78\\
\noalign{\smallskip}\hdashline[0.8pt/1pt]\noalign{\smallskip}
Baseline & 2 - 2 - 2 & 48.41 & 14.88\\
\noalign{\smallskip}
\textbf{\ours} & 2 - 2 - 2  & 30.71 & 88.38 \\
\noalign{\smallskip}\hline\noalign{\smallskip}
Baseline & 1 - 1 - 8 & 62.0 & 128.0 \\
\noalign{\smallskip}
\textbf{\ours} & 1 - 1 - 8  & 31.57 & 97.08\\
\noalign{\smallskip}\hline\noalign{\smallskip}
Baseline & 1 - 1 - 1 & 62.0 & 128.0 \\
\noalign{\smallskip}
\textbf{\ours} & 1 - 1 - 1  & 29.81 & 67.51\\
\noalign{\smallskip}\hline
\end{tabular}}}
\end{table}

\subsection{Ablations}
\label{sec:experiment_ablation}
As stated earlier, our main proposed modeling improvement is a combination of two methods: statistics-based quantization for the weights, and learning-based quantization for the activations.  We ablate the contribution of these methods and present the results in Table \ref{tab:ablation}.

The results clearly show that while each method can give moderate gains by itself over the baseline, these improvements are not sufficient by themselves to produce meaningful results.  None of the ablated models can achieve an R2 score above 1.5.  It's only the \emph{combination} of the two, which together stabilize the training and result in good convergence for fully ternary and binary models.

\begin{figure*}[t!]
    \centering
    \includegraphics[width=0.92\linewidth]{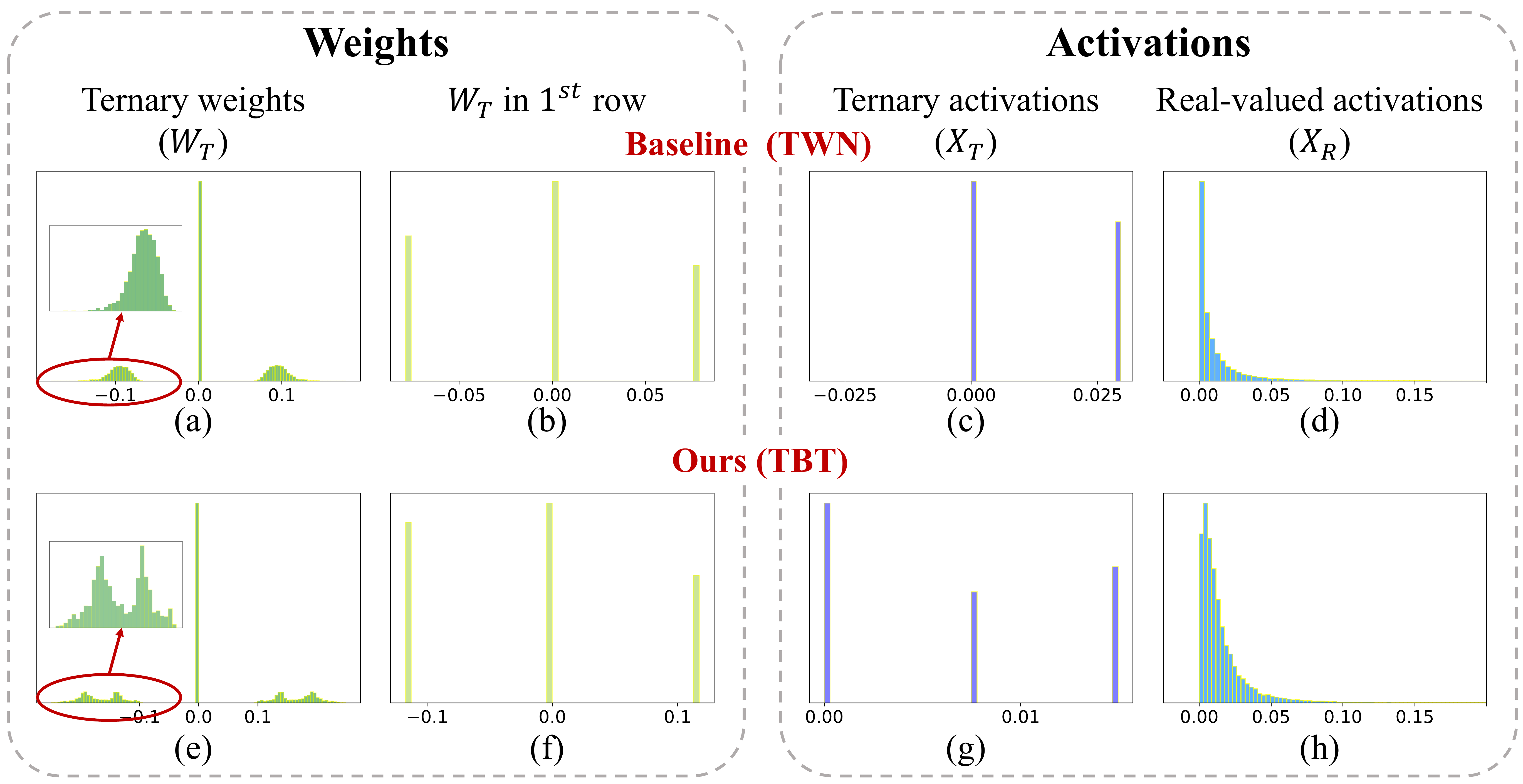}
    \caption{Weight and activation histogram comparison between the baseline TWN method and \ours{} method for ternarizing BART model on CNN/DailyMail benchmark. The weights are taken from the fully-connected layer of the value matrix in $1^{st}$ self-attention block in the decoder and activations are the attention outputs of the same layer.}
    \label{fig:hist}
\end{figure*}

\subsection{Sequence length analysis}
\label{sec:length}
In language generation tasks, the error compounding issue in the recursive decoder generation process will largely amplify the quantization error or even lead to divergent results, and thus is an harsh factor to test the robustness of a quantization method. The average generated sequence length indicates whether the quantized model can overcome the compounding error and generate reasonable length of text.

In Table~\ref{tab:length} we compare the generated sequence length between the proposed method and the baseline method (\textit{i.e.}, TWN~\cite{li2016ternary} for ternary, BWN~\cite{courbariaux2016binarized} for binary). Our method successfully produces summarizations with comparable length as the full-precision model on XSUM benchmark, even when both weights and activations are binarized.

Compared to XSUM dataset, for which the document are summarized to only one sentence, CNN/DailyMail is more challenging because it allows longer summary. We can clearly see that, the text generate with our 8-bit activation models can maintain near the similar average length as the full-precision BART model, while the binary and ternary activation models deviate moderately. In contrast, the baseline method is only able to derive reasonable summarization with 2-bit weight 8-bit activations and fails at lower bit-width, showing the difficult natural of the language generation tasks.

\subsection{Visualization}
To further understand the effectiveness of the proposed method, we visualize weight and activation histograms in the BART model ternarized with the baseline method and the proposed method in Fig.~\ref{fig:hist}.

Both the baseline method and our method use per-row weight ternarization, and thus a tensor tensor will have $\#$row of scaling factors. As we can see in Fig.~\ref{fig:hist} (b) and (g), the proposed method allows the weights to be more evenly distributed in three ternarization levels, which can allow higher information entropy in quantized weights, as discussed in Sec.~\ref{sec:stats_weight}. Additionally, we calculate the quantized weight distribution entropy (i.e., Eq.~\ref{eq:entropy}) in $96$ fully-connected layers in the BART-base model and found that the proposed $\ours{}$ method achieves consistently higher entropy in quantized weights than the baseline method in all the layers. Further, an interesting phenomenon we can see in Fig.~\ref{fig:hist} (a) (e) is that ternary weights in a baseline model are very close to the Gaussian distribution, in contrast, weights ternarized with $\ours{}$ are capturing a more sophisticated distribution. This phenomenon implies that the proposed method helps the weights learn more informative patterns and thus better satisfy the high demand for language generation tasks.

For activation quantization, it is evident that the attention layer and the SoftMax output only contain the positive activations ($\mathbf{X}_\mathbf{R} \in \mathbb{R}_+$). If simply ternarized to $\{-\alpha, 0, \alpha\}$, the ternary activations will waste one representative level (Fig.~\ref{fig:hist}(d)) and therefore lead to lower accuracy. Instead, the proposed method uses a two-set ternarization method that ternarizes the non-negative activation layer ($\mathbf{X}_\mathbf{R} \in \mathbb{R}_+$) to $\{0, \alpha, 2\alpha\}$, and learns the scaling factor $\alpha$ to better fit the underlying real-valued distribution. This ternarization method greatly reduces information loss and enhances the final accuracy.

\section{Related Work}
Quantization has long been studied to make neural networks more efficient (see ~\citep{hubara2017quantized} for a survey).  Due to the popularity of BERT, numerous works have studied quantization for transformer models, starting with 8-bit quantization~\citep{zafrir2019q8bert, fan2020training}, and progressing to 4-bit~\citep{shen2020q,zadeh2020gobo}, ternary~\citep{TernaryBERT} and binary~\citet{bai2021binarybert, qin2021bibert, liu2022bit}.  All of these works have focused on the encoder-only setting.

In the generative setting, \citet{prato2019fully, behnke2021efficient} demonstrate quantized models for machine translation, and \citet{fan2020training,bai2021towards} for language modeling, though only for moderate quantization levels (4-8 bits).  Most recently, \citet{tao2022compression} and \citet{li2022dq} pushed weight quantization down to 2 bits (with 8-bit activation quantization) and evaluated on language modeling and summarization.  However, our method outperforms these works substantially, while also demonstrating accurate generative transformers with both weights and activations quantized to 2-bit and even 1-bit for the first time.

\section{Conclusion}
We have demonstrated high accuracy ternary and binary natural language generation models based on a pre-trained transformer encoder-decoder backbone.  Quantizing both the weights and the activations of the network allow these models to run on special-purpose hardware using binary and ternary arithmetic, which doesn't require multiplication modules.  Therefore our results promise multiple orders of magnitude gains in efficiency while running these models, and can drastically expand the use cases of such models beyond just high end gpu servers.  We are especially excited about the implications of our results for larger text generation models such as GPT-3~\citep{brown2020language}.  These models have both demonstrated impressive capabilities, while also presenting enormous scaling and computational challenges.  Low-bit quantization is a promising approach to mitigate some of these issues.  Whether our approach will scale to these models is an open problem and an exciting future research direction.

\section{Limitations}
We conduct experiments on public datasets of finite sentence length, while generalizability to extremely long sequences or even streaming data has not been verified. Furthermore, the generalizability of the proposed quantization method to other tasks, including computer vision or speech recognition, remains to be tested. In addition, binarization and ternarization require bit-packing to have actual memory savings and dedicated hardware support for real-time acceleration, which is more of a hardware implementation aspect and not studied in this paper.

\section{Ethics Statement}
We affirm that we contribute to society, avoid harm, and are honest and trustworthy. We respect previous work and appropriately cite the methods and datasets we are using. All data we use is public and no private data is involved. There is some potential risk if the translation technique is maliciously used by a third party and thus we are committed to maintaining the compression techniques we have developed and the general summarization/machine translation techniques used correctly without incurring any form of discrimination.

\bibliography{main}
\bibliographystyle{acl_natbib}

\end{document}